\pdfoutput=1

\documentclass[11pt]{article}

\usepackage[]{acl}

\usepackage{times}
\usepackage{latexsym}

\usepackage[T1]{fontenc}

\usepackage[utf8]{inputenc}

\usepackage{microtype}
\usepackage{graphicx}
\usepackage{booktabs}
\usepackage{comment}
\usepackage[labelsep=quad,indention=10pt]{subfig}
\usepackage{bm}
\usepackage{wrapfig}
\usepackage[htt]{hyphenat}

\newcommand\VisualSem{VisualSem}

\title{\VisualSem: A High-quality Knowledge Graph for Vision \& Language}

\author{Houda Alberts$^{1,}$\thanks{$\quad$Work initiated while in the University of Amsterdam during her MSc. research.}\ \ ~~Ningyuan (Teresa) Huang$^5$\ \ ~~Yash R. Deshpande$^2$\ \  ~~Yibo Liu$^2$\\\bf Kyunghyun Cho$^2$\ \ Clara Vania$^{4}$\ \ Iacer Calixto$^{2,3}$\\\\
$^1$Rond Consulting, NL \ \ 
$^2$New York University \ \ 
$^3$ILLC, University of Amsterdam\\
$^4$Amazon, UK \ \
$^5$Johns Hopkins University, USA.\\
\texttt{houda1996@hotmail.com,iacer.calixto@nyu.edu}
}

\begin{document}
\maketitle
\begin{abstract}
An exciting frontier in natural language understanding (NLU) and generation (NLG) calls for (vision-and-) language models that can efficiently access external structured knowledge repositories. However, many existing knowledge bases only cover limited domains, or suffer from noisy data, and most of all are typically hard to integrate into neural language pipelines. To fill this gap, we release \VisualSem{}: a high-quality knowledge graph (KG) which includes nodes with multilingual glosses, multiple illustrative images, and visually relevant relations. We also release a neural multi-modal retrieval model that can use images or sentences as inputs and retrieves entities in the KG. This multi-modal retrieval model can be integrated into any (neural network) model pipeline. We encourage the research community to use \VisualSem{} for data augmentation and/or as a source of grounding, among other possible uses. \VisualSem{} as well as the multi-modal retrieval models are publicly available and can be downloaded in this URL:
\url{https://github.com/iacercalixto/visualsem}.
\end{abstract}

\section{Introduction}

Publicly-available multilingual resources such as Wikipedia are crucial when used as knowledge bases in recent neural language models (LMs) that retrieve documents in order to solve complex tasks such as open-ended question answering \citep{pmlr-v119-guu20a,NEURIPS2020-RAG,NEURIPS2020-MARGE}.
However, Wikipedia's wealth of visual information is typically hard to use,\footnote{See for instance \url{https://en.wikipedia.org/wiki/Wikipedia:Images} and \url{https://commons.wikimedia.org/wiki/Main_Page}.} which prevents models from including rich \textit{multi-modal data}. Although a small number of structured knowledge graphs (KGs) with images exist and are publicly available, they either cover limited domains \citep{Ruobing-etal-2017-ikrl,moussellySergieh2018multimodal} or span multiple domains but with noisy images \citep{navigli2012babelnet}.
To facilitate further progress in this direction, we introduce a new resource to enable research on LMs that efficiently retrieve \textit{textual} and \textit{visual contextual information} from KGs, where this information can be used for grounding, data augmentation, among other uses.

In this paper, we introduce \VisualSem{}, a multilingual and multimodal KG with $\sim90$k nodes, $\sim1.3$M glosses and $\sim938$k images.
\VisualSem{}'s nodes denote concepts and named entities, include multiple curated illustrative images, as well as glosses in up to 14 diverse languages.
Typed semantic relations connect nodes in the graph, and nodes in \VisualSem{} are linked to Wikipedia articles, WordNet synsets, and (when available) high-quality images from ImageNet \citep{Deng-etal-2009-ImageNet}.  \VisualSem{} integrates seamlessly with existing resources.
Compared to existing multimodal KGs, \VisualSem{} includes data from different sources and thus it is also more diverse in terms of the domains.
We source the images in \VisualSem{} using BabelNet \citep{navigli2012babelnet}, a large multilingual and multimodal resource that semi-automatically aggregates information from many different sources.
We address the known issue of noisy images in BabelNet \citep{Collaetal2018,calabrese-etal-2020-fatality} by applying multiple filtering steps to ensure that we remove noise while maintaining breadth of coverage and image diversity.

We also release pre-trained models to retrieve entities from VisualSem using either images or sentences as queries in a $k$-nearest neighbor search. This effectively allows researchers to integrate entities and facts in VisualSem into their (neural) model pipelines.
Code to generate and download \VisualSem{}, as well as retrieval models, is publicly available.\footnote{
\url{https://github.com/iacercalixto/visualsem}.
}

\begin{figure*}[t!]
    \centering
    \subfloat[Node ``Noah''.]{%
        \includegraphics[width=.33\textwidth]{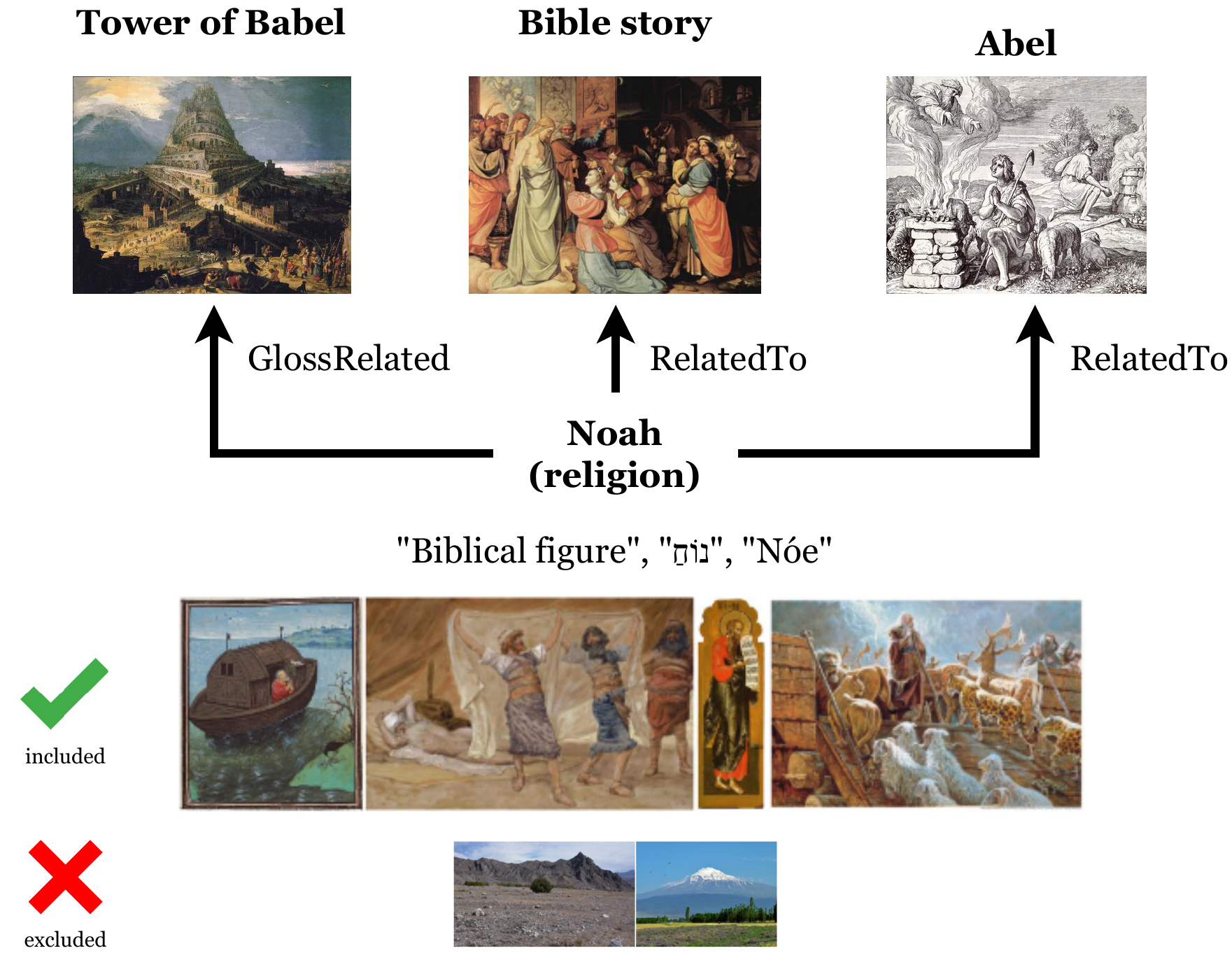}%
        \label{fig:visualsem_example_noah}%
    }\hfill
    \subfloat[Node ``Air quality index''.]{%
        \includegraphics[width=.33\textwidth]{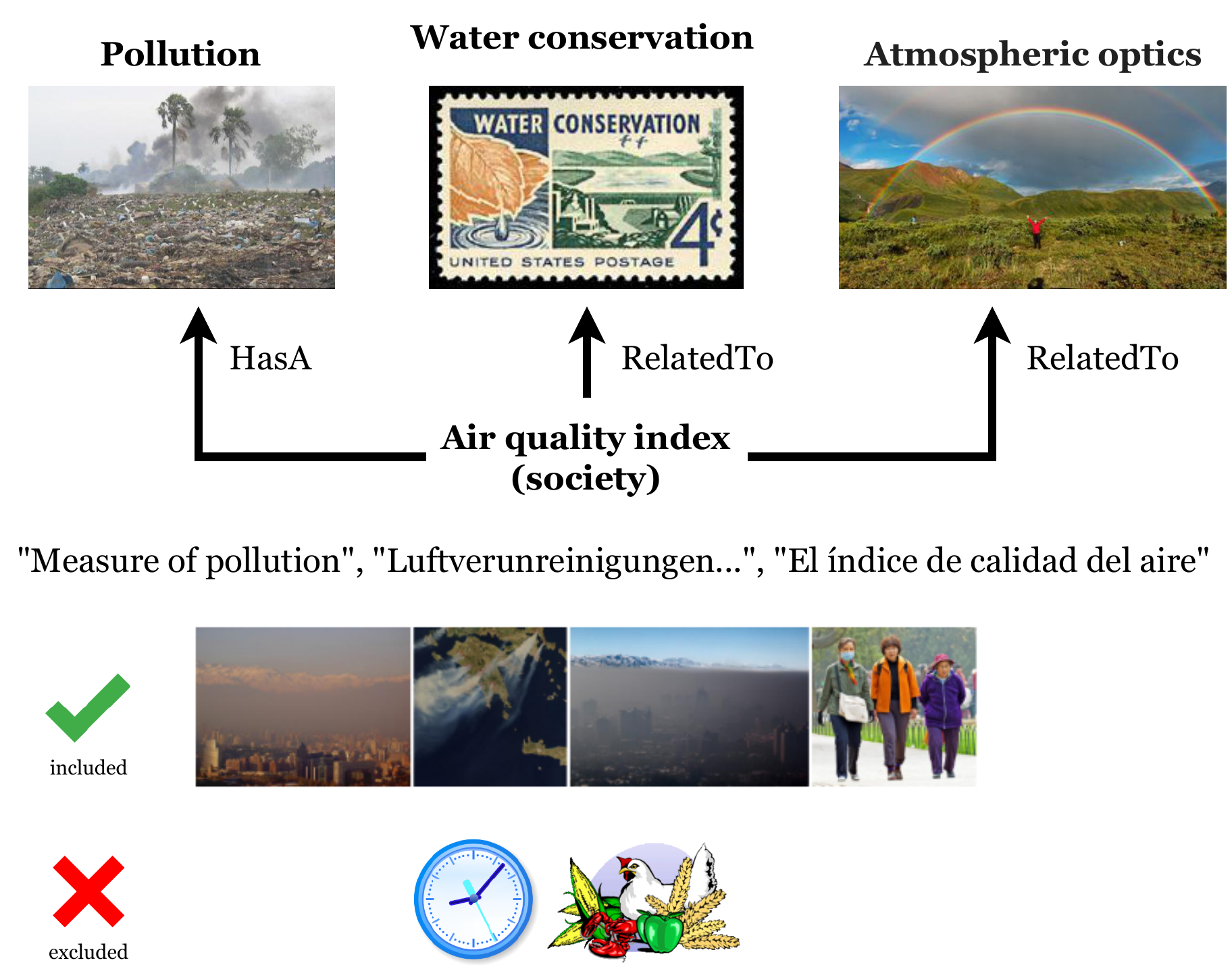}%
        \label{fig:visualsem_example_air}%
    }\hfill%
    \subfloat[Node ``Ronaldo''.]{%
        \includegraphics[width=.33\textwidth]{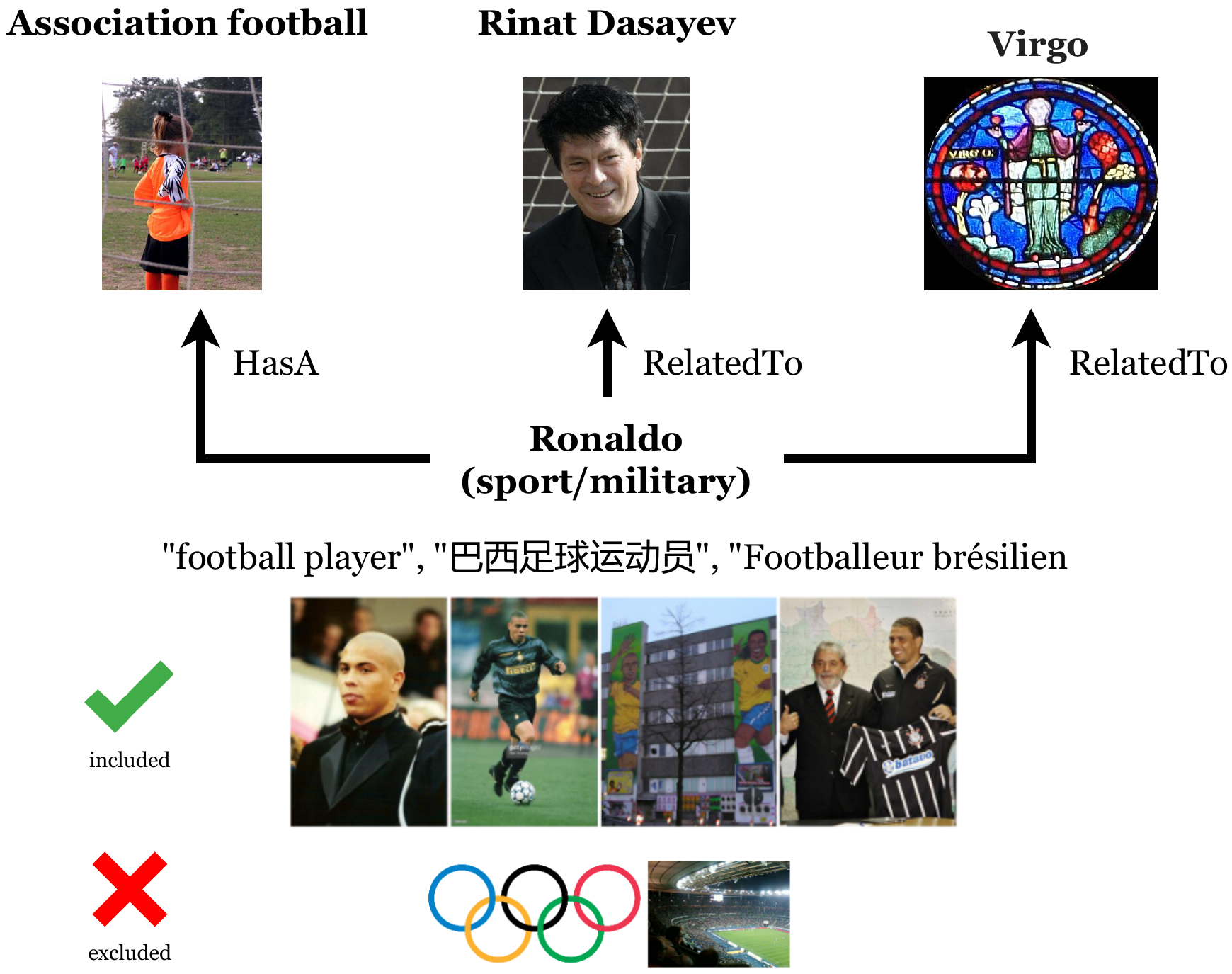}%
        \label{fig:visualsem_example_ronaldo}%
    }
    \caption{Example nodes in \VisualSem{}, some of their glosses and images, and how they relate to other nodes. We also show examples of images we collected for the nodes that were filtered out and that were kept in following our data collection pipeline (Section \ref{sec:data_collection}).
    }
    \label{fig:visualsem_example_nodes}
\end{figure*}

Our main contributions are:
\begin{itemize}
    \item We introduce \VisualSem{}, a multi-modal knowledge graph designed to be used in vision and language research that integrates textual descriptions in up to $14$ languages and images from curated sources.
    \item We build an image filtering pipeline to obtain a clean set of images associated to each concept in \VisualSem{}.
    \item We provide an open source code base one can use to download and generate the KG, as well as multi-modal retrieval models that retrieve entities from the KG given images and sentences.
\end{itemize}

This paper is organised as follows.
In Section~\ref{sec:approach} we explain how we build \VisualSem{} and provide details on its data collection pipeline.
In Section~\ref{sec:statistics_and_analysis} we include dataset statistics and also analyse \VisualSem{}'s  content and structure qualitatively.
In Section~\ref{sec:knowledge_base_retrieval}, we describe the multi-modal retrieval models that use sentences and images to retrieve entities from the KG.
In Section~\ref{sec:related} we discuss relevant related work, including existing multi-modal knowledge bases and how they compare to our work.
Finally, in Section~\ref{sec:conclusions} we discuss our main findings as well as provide avenues for future work.
\section{Approach}
\label{sec:approach}

\VisualSem{} is a multilingual and multimodal knowledge graph consisting of $89,896$ unique nodes and $1,481,007$ facts,
where each fact is a tuple \texttt{<}$n_i$, $r_j$, $n_k$\texttt{>} that denotes that node $n_i$ is connected to node $n_k$ via relation $r_j$.
Each node $n$ denotes a \textit{concept}, such as a WordNet synset or Wikipedia article (e.g., see example nodes in Figures \ref{fig:visualsem_example_noah}, \ref{fig:visualsem_example_air}, and \ref{fig:visualsem_example_ronaldo}).
Relations $r_j$ can take one of $13$ different semantic types.
We select \VisualSem{}'s nodes and relations carefully:
we include nodes that denote concepts and named entities with a strong visual component, and relations that encode relevant visual knowledge and/or knowledge relevant to solving tasks that require visual understanding (discussed next).
We extract nodes, relations, and images in \VisualSem{} using BabelNet v4.0 \citep{navigli2012babelnet}, a large multilingual KB that consolidates data from multiple knowledge graphs into one single graph.\footnote{\url{https://babelnet.org}}

\paragraph{Relation types}
We follow previous work to choose relation types in \VisualSem{}.
\citet{Cuietal2018} propose to use $15$ relation types with a strong visual component and that therefore are likely to help in vision and language tasks: \textit{is-a}, \textit{has-part}, \textit{related-to}, \textit{used-for}, \textit{used-by}, \textit{subject-of}, \textit{depicts}, \textit{receives-action}, \textit{made-of}, \textit{has-property}, \textit{also-see}, \textit{gloss-related}, \textit{synonym}, \textit{part-of}, and \textit{located-at}.
We use $13$ out of the $15$ proposed relation types, since we do not have examples with \textit{depicts} and \textit{also-see} in the nodes we select.
We describe the data collection procedure next.

\subsection{Data collection}\label{sec:data_collection}

We start by choosing a set of \textit{seed} nodes we can guarantee is high-quality, well-curated, and visually relevant.
We therefore use the BabelNet API\footnote{\url{https://babelnet.org/guide}} to get synsets corresponding to the $1,000$ ImageNet classes
used in the ILSVRC image classification competition~\citep{ILSVRC2015} as our initial seed nodes.\footnote{\url{https://image-net.org/challenges/LSVRC/2014/browse-synsets}}
We call these nodes our initial \textit{node pool}.
We choose this initial node pool since ImageNet already provides high-quality images associated to these nodes.
We then follow an iterative procedure where in each step we add nodes to the node pool, until we reach the end of the algorithm.
Specifically, in the first step, the node pool includes the $1,000$ seed nodes; in the second step, it will include the $1,000$ seed nodes plus the additional nodes gathered in the first step; and so on, until we reach $N$ nodes ($N=90,000$).
This procedure works by iterating over the following steps:
\begin{enumerate}
    \item {\bf Retrieve neighbors}: retrieve neighbor nodes for each node in the node pool using the BabelNet API;
    \item {\bf Validate images}:
    validate linked images and optionally remove those that do not meet certain quality criteria;
    \item {\bf Filter nodes}: filter out nodes that do not meet inclusion criteria;
    \item {\bf Update pool}: accept top-$k$ nodes among remaining nodes after sorting nodes according to their features.
\end{enumerate}

\paragraph{Retrieve neighbors} In this step we collect all first-degree neighbors for each node in the node pool, and remove any duplicate node retrieved if any exists.
The neighbors are retrieved using the BabelNet v4.0 API \citep{navigli2012babelnet} and can include nodes from multiple sources, such as multilingual Wikipedias, multilingual WordNets, among others.
First-degree neighbors $n_k$ of node $n_i$ are those nodes directly connected via a relation which type $r_j$ is one of the 13 relation types, i.e., \texttt{<}$n_i$, $r_j$, $n_k =$  $?$\texttt{>}.
In Table \ref{tab:visualsem_relations} we show a list of relation types in BabelNet, and how these were merged into the 13 types used in \VisualSem{}.

\begin{table}[t!]
    \centering\small
    \begin{tabular}{rr}
        \toprule
        BabelNet & \VisualSem{} \\
        \midrule
        \texttt{is-a}, \texttt{is\_a} & \texttt{is-a} \\
        \texttt{has-part}, \texttt{has\_part} & \texttt{has-part} \\
        \texttt{related} & \texttt{related-to} \\
        \texttt{use} & \texttt{used-for} \\
        \texttt{used-by}, \texttt{used\_by} & \texttt{used-by} \\
        \texttt{subject-of}, \texttt{subject\_of} & \texttt{subject-of} \\
        \texttt{interaction} & \texttt{receives-action} \\
        \texttt{oath-made-by} & \texttt{made-of} \\
        \texttt{has\_*} & \texttt{has-property} \\
        \texttt{gloss-related} & \texttt{gloss-related} \\
        \texttt{taxon-synonym} & \texttt{synonym} \\
        \texttt{part-of}, \texttt{part\_of} & \texttt{part-of} \\
        \texttt{location}, \texttt{located\_*} & \texttt{located-at} \\
        \bottomrule
    \end{tabular}
    \caption{Relation types in BabelNet and their corresponding types in \VisualSem{}. Asterisks (\texttt{*}) can match any number of characters.}
    \label{tab:visualsem_relations}
\end{table}

\begin{table*}[t!]
    \small
    \centering
    \resizebox{\textwidth}{!}{
    \begin{tabular}{ll rrrrrrrrr}
        \toprule
        &
        \# langs.
        & \# nodes & \# rel. types & \# glosses & \# images & \# train & \# valid & \# test & sources \\
        \midrule
        {\bf WN9-IMG}$^\dagger$ &
        1
        & $6,555$ & $9$ & N/A
        & $65,550$ & $11,741$ & $1,337$ & $1,319$ & WordNet \\
        {\bf FB15-IMG}$^\ddagger$ &
        1
        & $11,757$ & $1,231$ & N/A & $107,570$ & $285,850$ & $29,580$ & $34,863$ & Freebase \\
        \cmidrule{2-10}
        {\bf VisualSem} &
        14
        &
        $89,896$ & $13$ & $1,342,764$ & $938,100$ & $1,441,007$ & $20,000$ & $20,000$ & Multiple$^\star$ \\
        \bottomrule
    \end{tabular}
    }
    \caption{\VisualSem{} KG statistics.
    $^\star$Multiple sources include Wikipedia, WordNet, ImageNet, among others.
    $^\dagger$\citet{Ruobing-etal-2017-ikrl}.
    $^\ddagger$\citet{moussellySergieh2018multimodal}.
    }
    \label{tab:data_statistics}
\end{table*}

\paragraph{Validate images + Node filtering + Update pool}
We collect images available for all first-degree neighbors retrieved in the previous step, and apply \textit{four filters} to the available images: {\bf i)} we check if images are valid files, {\bf ii)} we remove duplicate or near-duplicate images, {\bf iii)} we train a binary classifier to remove non-photographic images, and {\bf iv)} we use OpenAI's CLIP \citep{radford2021learning} to remove images that do not minimally match \textit{any} of the node's glosses.
At the end of each iteration, we require that each node has at least one image associated to it, and that it contains relations with other nodes with a minimum of two different relation types.
If nodes do not satisfy these criteria, they are filtered out.

In more detail: {\bf i)} From the total amount of image files we downloaded, an average of $\sim6.3\%$ were invalid image files and thus removed, e.g., the downloaded file was in fact an audio file.
{\bf ii)} We find many images that are \mbox{(near-)duplicates} across different synsets, in which case we remove those duplicates using SHA1 hashing \citep{eastlake-jones:2001sha1}.
{\bf iii)} We use a binary classifier to validate images associated to each node in the pool of retrieved neighbors following the simple procedure described in \citet{ImagiFilter2020}.
We use their ResNet-based coarse-grained binary classification model to further filter out undesirable images.
In short, the image classification model is trained on a total of $6,000$ randomly sampled images (mostly sourced from Wikipedia and ImageNet) where 50\% are good quality/photographic images, and the remaining are non-photographic/undesirable ones.
We denote undesirable images as unclear/blurred/dark or low-quality images, and non-photographic images such as hand drawings, sketches, maps, icons, flags, graphs and other rendered images.
Please refer to \citet{ImagiFilter2020} for more details on the image classification model training and evaluation.
We apply this step and remove images flagged with this binary classifier, since many images we obtain in practice are very noisy.
{\bf iv)} Finally, in our last filter in the image filtering process we use the pre-trained CLIP model.
CLIP has one text and one image encoder and is trained to maximize the dot-product of correct image-sentence pairs, and at the same time minimize that of random image-sentence pairs.
We encode the $k$ English glosses $g_{i,k}$ and the $l$ images $a_{i,l}$ available for node $n_i$ with CLIP's text and image encoder, respectively.
We then compute the dot-product between each image $a_{i,l}$ and each gloss $g_{i,k}$, keeping only the images that had at least one dot-product greater than $0.5$ with one of the English glosses. We choose the value of $0.5$ empirically by manually checking the quality of the image-gloss matches.
We note that the motivation for keeping only images that match well with at least one gloss is the fact that the available glosses are by design descriptive of the node, and by making sure images align well with at least one of the available glosses we can make sure we filter out noisy and unrelated images (e.g., please refer to examples in Appendix~\ref{sec:appendix_examples} for more details).

Step (i) reduces the number of images from $\sim5.6$ to $\sim5.3$ million, step (ii) brings the number of images from $\sim5.3$ to $\sim2.1$ million, step (iii) further reduces images from $\sim2.1$ to $\sim1.5$ million, and finally step (iv) brings us to the final number of images in \VisualSem{}, $938,100$.

After filtering out undesirable images in the retrieved neighbors, we discard any nodes in the neighborhood that do not have at least one associated image.
Finally, for each node in the initial node pool we add the top-$k$ neighbour nodes ($k=10$) to the pool and repeat.
We prioritize visually relevant nodes with a larger number of images, nodes that include as many diverse relation types as possible, and especially nodes with less frequent relation types.

\begin{figure*}[t!]
    \centering
    \subfloat[Number of images per node.]{%
        \includegraphics[width=.33\textwidth]{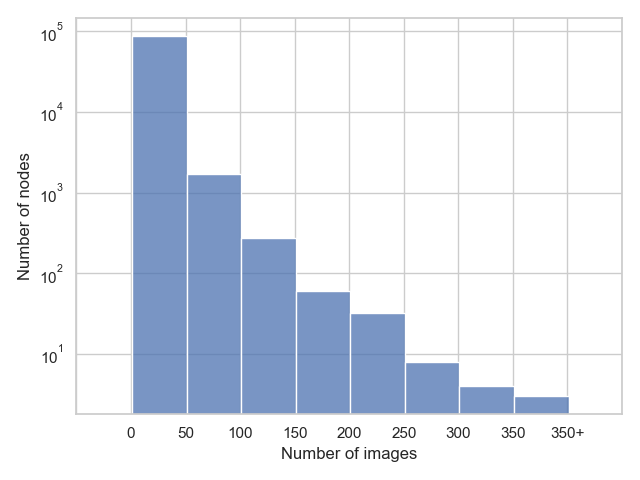}%
        \label{fig:images_per_node}%
    }\hfill 
    \subfloat[Number of instances or facts (i.e., tuples \texttt{<}$n_i$, $r_j$, $n_k$\texttt{>}) per relation type.]{%
        \includegraphics[width=.33\textwidth]{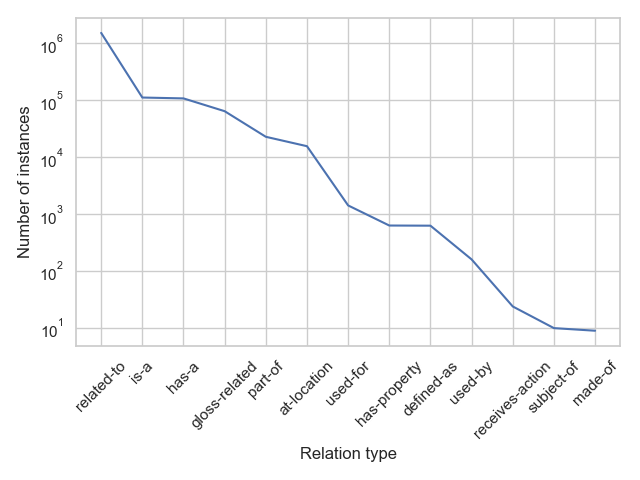}%
        \label{fig:instances_per_relation_type}%
    }\hfill 
    \subfloat[Number of nodes with at least one gloss in each language.]{%
        \includegraphics[width=.33\textwidth]{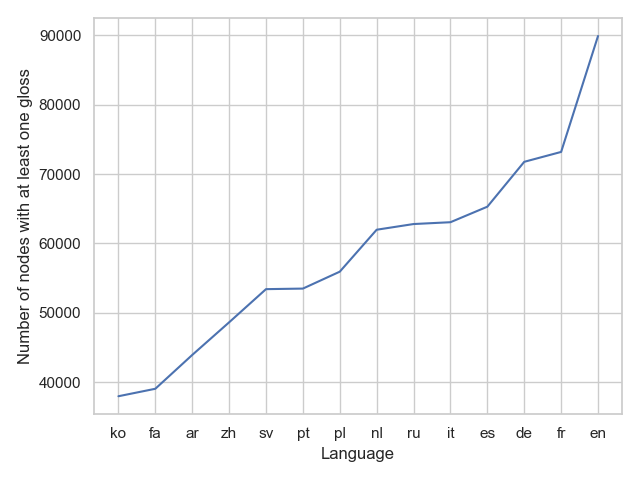}%
        \label{fig:nodes_with_at_least_one_gloss_per_language}%
    }
    \caption{\VisualSem{} data statistics.
    }
    \label{fig:stats_visualization}
\end{figure*}

\section{Data Statistics and Analysis}
\label{sec:statistics_and_analysis}

In this Section, we explore the structure and content of \VisualSem.
In Section \ref{sec:statistics}, we report relevant statistics, as well as show some examples of nodes and relations in it.
In Section \ref{sec:topic_models}, we provide a more qualitative analysis using unsupervised topic models to induce topic distributions for \VisualSem{}.

\subsection{Data Statistics}
\label{sec:statistics}

In Table \ref{tab:data_statistics}, we show statistics comparing \VisualSem{} and multimodal knowledge bases WN9-IMG and FB15-IMG.

\paragraph{Glosses.} \VisualSem{} has a total of $1,342,764$ glosses in $14$ different languages: Arabic, Chinese, Dutch, English, Farsi, French, German, Italian, Korean, Polish, Portuguese, Russian, Spanish, and Swedish.
The languages were chosen to be representative of diverse linguistic families and to include many different scripts, while at the same time covering a high number of nodes.
Nodes have on average $14.9$ glosses across all $14$ languages, and in Figure \ref{fig:nodes_with_at_least_one_gloss_per_language} we show the number of nodes that have at least one gloss in each language.
Nodes with English (Korean) glosses have the highest (lowest) coverage: $89,896$ nodes have at least one English and $37,970$ at least one Korean gloss.
\textbf{Example:} Node \textit{Ronaldo (Brazilian footballer)} has four glosses in English, and one of these glosses is \textit{Ronaldo Luís Nazário de Lima, commonly known as Ronaldo, is a retired Brazilian professional footballer who played as a striker}\footnote{The same node also has glosses in French (3), German (2), Spanish (4), Italian (4), Russian (1), Dutch (3), Polish (1), Portuguese (3), Swedish (3), Mandarin (1), Arabic (1), Farsi (2), and Korean (1).} (see Figure \ref{fig:visualsem_example_ronaldo}).

\paragraph{Images.}
As shown in Table \ref{tab:data_statistics}, \VisualSem{} has a total of $938,100$ unique images.
On average, there are $10.4$ images per node---similarly to WN9-IMG and FB15-IMG datasets---and in Figure \ref{fig:images_per_node} we show the distribution of images per node.
The node with the greatest number of images is the concept \textit{Russian culture} with $848$ images.\footnote{It has BabelNet id \texttt{bn:01286889n} and is linked to Wikipedia article \url{https://en.wikipedia.org/wiki/Russian_culture}.}

\paragraph{Relations.}
\VisualSem{} has $13$ different relation types, and all its relations are typed.
Relation types include: \textit{is-a}, \textit{has-part}, \textit{related-to}, \textit{used-for}, \textit{used-by}, \textit{subject-of}, \textit{receives-action}, \textit{made-of}, \textit{has-property}, \textit{gloss-related}, \textit{synonym}, \textit{part-of}, and \textit{located-at}.
In Table \ref{tab:visualsem_relations} we show the mapping from relations in BabelNet to the corresponding types in \VisualSem{}.
\textbf{Example:} In Figure \ref{fig:visualsem_example_nodes} we show three example nodes and how they relate to other nodes in the KG.
We note that most connections have the \textit{related-to} relation type.
In Figure \ref{fig:instances_per_relation_type}, we show the number of instances (i.e., tuples \texttt{<}$n_i$, $r_j$, $n_k$\texttt{>}) per relation type.
There is a very large imbalance, with the type \textit{related-to} accounting for about $82\%$ of all existing relations in the KB.
We note that had we not adopted measures to alleviate this issue in our data collection pipeline---e.g., prioritizing nodes that have less frequent relation types when adding to the node pool---this imbalance would have been much greater (see Section \ref{sec:data_collection}).

\subsection{Topic Models}\label{sec:topic_models}
To further analyze the quality of our KG, we try to understand what topics are salient in \VisualSem{}.
We train a neural topic model with $20$ latent topics using Embedded Topic Model \citep[ETM;][]{dieng-etal-2020-topic}, which is an extension of Latent Dirichlet Allocation \cite[LDA;][]{Blei-etal-2003-LDA} that use pre-trained word embeddings.
We let each node be a document, and each document's content be the combination of all its English glosses.
In Table~\ref{tab:lda_results}, we show the six most representative words per topic after stop-word removal.
The topics covered are varied and can be clustered in broad domains such as `sciences' (e.g., physics, chemistry, biology, space), `society' (e.g., geography, politics, occupations), `culture' (e.g., religion, history, food, fashion), among others (e.g., material, city). There is a trend towards representing factual knowledge, which is expected since glosses by definition describe facts about nodes.
Concepts well covered in Wikipedia are also well covered in \VisualSem{}.

\begin{table*}[ht!]
    \centering
    \resizebox{\textwidth}{!}{
    \begin{tabular}{cccccccccc}
        \toprule
        {\bf 1. space} & {\bf 2. occupation} & {\bf 3. politics} & {\bf 4. chemistry} & {\bf 5. food} & {\bf 6. occupation} & {\bf 7. society} & {\bf 8. history} & {\bf 9. religion} & {\bf 10. country}\\
        \midrule 
        \texttt{planet} & \texttt{dance} & \texttt{party} & \texttt{gas} & \texttt{meat} & \texttt{actor} & \texttt{school} & \texttt{rome} & \texttt{religion} & \texttt{commune} \\ 
        \texttt{constellation} & \texttt{physicist} & \texttt{rank} & \texttt{formula} & \texttt{bread} & \texttt{composer} & \texttt{institution} & \texttt{emperor} & \texttt{directed} & \texttt{autonomous} \\ 
        \texttt{boat} & \texttt{painting} & \texttt{officer} & \texttt{atomic} & \texttt{cheese} & \texttt{band} & \texttt{economic} & \texttt{ireland} & \texttt{jesus} & \texttt{saxony} \\ 
        \texttt{spacecraft} & \texttt{mathematician} & \texttt{politician} & \texttt{acid} & \texttt{sauce} & \texttt{painter} & \texttt{education} & \texttt{pope} & \texttt{jewish} & \texttt{philippine} \\ 
        \texttt{moon} & \texttt{philosopher} & \texttt{minister} & \texttt{iron} & \texttt{vegetable} & \texttt{singer} & \texttt{agency} & \texttt{dynasty} & \texttt{bible} & \texttt{indonesia} \\ 
        \texttt{mar} & \texttt{scientist} & \texttt{currency} & \texttt{solid} & \texttt{rice} & \texttt{musician} & \texttt{society} & \texttt{egypt} & \texttt{goddess} & \texttt{finland} \\
        \midrule\midrule
        {\bf 11. sports/military} & {\bf 12. technology} & {\bf 13. mixed} & {\bf 14. physics} & {\bf 15. geography} & {\bf 16. medicine} & 
        {\bf 17. material} & {\bf 18. fashion} & {\bf 19. biology } & {\bf 20. city}\\
        \midrule 
        \texttt{football} & \texttt{electrical} & \texttt{horse} & \texttt{image} & \texttt{bridge} & \texttt{blood} & \texttt{garment} & \texttt{hair} & \texttt{cat} & \texttt{museum} \\ 
        \texttt{tank} & \texttt{data} & \texttt{ice} & \texttt{measure} & \texttt{switzerland} & \texttt{bone} & \texttt{clothing} & \texttt{fabric} & \texttt{shark} & \texttt{street} \\ 
        \texttt{rifle} & \texttt{storage} & \texttt{wall} & \texttt{motion} & \texttt{wine} & \texttt{tissue} & \texttt{dress} & \texttt{soil} & \texttt{temperate} & \texttt{metro} \\ 
        \texttt{team} & \texttt{electronic} & \texttt{blade} & \texttt{energy} & \texttt{valley} & \texttt{muscle} & \texttt{skirt} & \texttt{colour} & \texttt{subfamily} & \texttt{stockholm} \\ 
        \texttt{carrier} & \texttt{signal} & \texttt{tool} & \texttt{wave} & \texttt{canton} & \texttt{organism} & \texttt{flag} & \texttt{cloth} & \texttt{beetle} & \texttt{tokyo} \\ 
        \texttt{stadium} & \texttt{card} & \texttt{stone} & \texttt{radiation} & \texttt{archipelago} & \texttt{organ} & \texttt{garden} & \texttt{hat} & \texttt{grass} & \texttt{korea} \\ 
        \bottomrule
    \end{tabular}
    }
    \caption{Topics induced using the Embedded Topic Model on \VisualSem{} English glosses (labels in bold are assigned manually).}
    \label{tab:lda_results}
\end{table*}

\begin{figure}[t!]
    \centering
    \includegraphics[width=\linewidth]{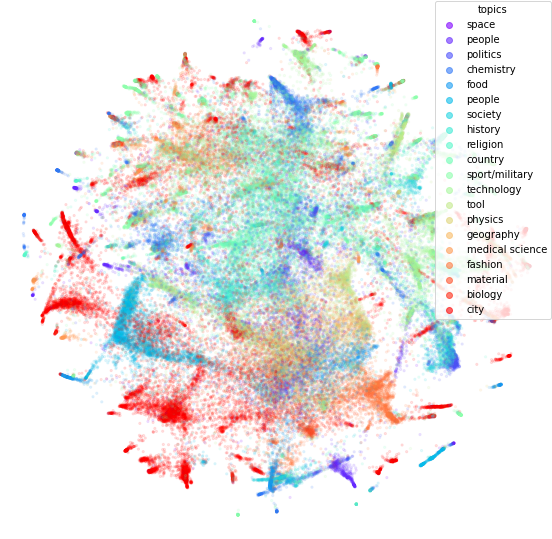}
    \caption{T-SNE plot for node embeddings where each node is represented as the average of its gloss embeddings. Topic assignments are used to colorise node embeddings, and topics are computed with the topic model described in Section \ref{sec:topic_models} and Table \ref{tab:lda_results}.}
    \label{fig:gloss features}
\end{figure}

\subsection{Node visualisation via its glosses}\label{sec:visualise_gloss}

We now visualise nodes with the t-SNE algorithm \citep{vanDerMaaten2008}.
We embed glosses as the average word embedding for each word in the gloss computed with multilingual fastText \citep{bojanowski2016enriching}.
Similarly, we set each node $n_i$'s representation $\bm{n}_i$ as the average of all its gloss embeddings, across all languages.
We use the ETM topic model trained in Section \ref{sec:topic_models} to assign a topic for each node in \VisualSem{} (i.e., the highest probability topic induced in the topic model), and colorise the node according to its predicted topic.
We apply t-SNE to the node embeddings $\bm{n}_i$ and show the plot in Figure \ref{fig:gloss features}.\footnote{We use the following hyperparameters with t-SNE: perplexity is set to 70, and number of iterations to 5000.}
We highlight that nodes cluster well according to their predicted topic, and note that even though there clearly is some structure in the space induced by the KG glosses, finding hard boundaries across topics does not seem straightforward (at least for the majority of nodes closer to the center of the plot).

\section{Entity Retrieval}\label{sec:knowledge_base_retrieval}

In this section, we describe \textit{retrieval models} that retrieve entities in the KB given a sentence or an image.
Our use-case includes tasks with possibly multi-modal inputs---i.e., either a sentence, an image, or both---where we use these inputs to retrieve relevant entities from \VisualSem{}.
The idea is to ground the task inputs or to augment the available data to train the task model.
This task could be text-only, such as named entity recognition or machine translation, or multi-modal, such as visual question answering or image captioning. We do not include experiments on specific tasks but instead leave that investigation for future work.

We use a standard \textit{retrieval as ranking} framework where the model is trained to rank nodes in \VisualSem{} given an input, and we use the top-$k$ ranked nodes as the results.
These retrieval modules allow for an easy integration of \VisualSem{}'s entities, glosses, and images into neural pipelines.

\begin{table}[t!]
    \centering
    \begin{tabular}{lrrr}
        \toprule
        & \# train & \# valid & \# test \\
        \midrule
        Glosses &
        $1,286,764$ & 
        $28,000$ & $28,000$ \\
        Images & 
        $898,100$ & 
        $20,000$ & $20,000$ \\
        \bottomrule
    \end{tabular}
    \caption{\VisualSem{} gloss and image splits. Gloss validation and test splits include $2,000$ entries for each of the $14$ languages in \VisualSem{}.}
    \label{tab:visualsem_gloss_and_image_splits}
\end{table}

We frame entity retrieval using sentences (henceforth \textit{sentence retrieval}, Section \ref{sec:sentence_retrieval}) as a sentence-to-gloss ranking problem, and use glosses available in all languages in the model.
Given an input sentence, we transform the resulting ranking over glosses into a ranking over nodes (i.e., by retrieving the nodes these glosses are linked to from their 1-to-1 mappings).
We similarly frame retrieval using images (henceforth \textit{image retrieval}, Section \ref{sec:image_retrieval}) as an image-to-gloss ranking problem, and use all English glosses available for retrieval.

Experiments in this section use the gloss and image splits in Table \ref{tab:visualsem_gloss_and_image_splits}.
Training, validation and test splits for glosses and images are chosen randomly in order to facilitate their use by the research community in future experiments.
Gloss validation and test splits are balanced regarding different languages, and each split includes $2,000$ examples for each of the $14$ languages in \VisualSem{}.

\begin{table}[t!]
    \centering
    \begin{tabular}{rrrrr}
        \toprule
        & \multicolumn{3}{c}{\bf Hits@k} & \multicolumn{1}{c}{\bf Rank} \\
        & \bf 1  $\uparrow$ & \bf 3 $\uparrow$ & \bf 10 $\uparrow$ & \multicolumn{1}{c}{\bf mean (std) $\downarrow$} \\
        \midrule
        ar & 37.7 & 48.9 & 58.6 & 2,572 (18,369)\\
        de & 48.9 & 58.3 & 66.7 & 1,590 (13,801)\\
        en & 56.1 & 64.0 & 73.0  & 15,156 (133,161)\\
        es & 60.5 & 69.3 & 76.4  & \underline{693 (6,234)}\\
        fr & 53.4 & 62.3 & 70.1  & 1,967 (20,850)\\
        it & 57.7 & 66.1 & 73.2  & 1,248 (16,216)\\
        ko & 44.7 & 56.8 & 66.8  & 1,488 (19,586)\\
        nl & 46.2 & 54.8 & 62.6  & 3,110 (31,413)\\
        pt & \bf 73.1 & \bf 79.5 & \bf 83.8  & 1,646 (34,586)\\
        ru & 28.9 & 36.4 & 42.8  & 16,043 (55,115)\\
        zh & \underline{62.9} & \underline{73.3} & \underline{81.1}  & 1,691 (26,218)\\
        fa & 38.6 & 49.8 & 60.1  & 1,829 (9,089)\\
        pl & 49.2 & 58.0 & 66.8  & 3,803 (25,605)\\
        sv & 61.7 & 71.4 & 78.7  & \bf 656 (6,865)\\
        \bf avg. & 51.4 & 60.6 & 68.6  & 3,821 (43,430)\\
        \bottomrule
    \end{tabular}
    \caption{Sentence retrieval results on \VisualSem{} test glosses. We report Hits@$k$, which is the percentage of time the correct node is retrieved in the top-$k$ results (higher is better) and the mean rank of the correct node (lower is better). According to Hits@$k$, Portuguese and Chinese have the best retrieval results, whereas worst results are obtained for Russian, Arabic, and Farsi. Best mean ranks are obtained with Spanish and Slovenian queries, and worst with Russian and English.}
    \label{tab:sentence_retrieval_results}
\end{table}

\subsection{Sentence retrieval}\label{sec:sentence_retrieval}
\VisualSem{} has $N = 89,896$ nodes with glosses in up to 14 languages.
We encode glosses using Sentence BERT \cite[SBERT;][]{reimers-2019-sentence-bert}, which is a family of models based on bi-encoder architectures trained on the task of sentence similarity that have strong performance when used for retrieval/indexing.
SBERT models are by default trained on English-only sentences to map semantically similar sentences to points close in the embedding space.
We use a multilingual SBERT model trained using knowledge distillation to transfer knowledge from English to other languages, more specifically the \texttt{paraphrase-multilingual-mpnet-base-v2} model \citep{reimers-2020-multilingual-sentence-bert} trained on over 50 languages. We select the model from all publicly available models according to validation performance.\footnote{Models available in \url{https://www.sbert.net/docs/pretrained_models.html\#multi-lingual-models}.}

Let $g_{i,j}$ be the $j$-th gloss in  the set of glosses associated to node $i$, and $\bm{g}_{i,j}$ be gloss $g_{i,j}$'s 512-dimensional vector representation computed with SBERT.
We implement the sentence retrieval using $k$-nearest neighbour ($k$-NN).
We directly rank all glosses in the split given the query according to their cosine similarity, and use the node associated to the gloss with the highest cosine similarity as the retrieved node.

\paragraph{Evaluation and Discussion}
We use the SBERT model \texttt{paraphrase-multilingual-mpnet-base-v2} \citep{reimers-2019-sentence-bert,reimers-2020-multilingual-sentence-bert}, trained on over 50 languages using knowledge distillation on the task of sentence similarity.
All languages covered in \VisualSem{} are supported by the model.
We directly use the $28,000$ held-out test glosses ($2$k per language) for model evaluation and show the results in Table \ref{tab:sentence_retrieval_results}.
We note that retrieval results are in general very good according to Hits@$k$.
Portuguese and Chinese queries have the best sentence retrieval performance according to this metric, whereas querying the model using Russian, Arabic, and Farsi has the worse performance.
We note that mean ranks show high variances, which indicate that retrieved nodes could be noisy despite the good Hits@$k$ scores.
Surprisingly, mean ranks obtained with English and Russian queries are the highest (i.e., the lower the mean ranks, the better).
We recommend that users use the sentence retrieval model with care when dealing with any of the low-quality languages.

\subsection{Image retrieval}\label{sec:image_retrieval}
CLIP \citep{radford2021learning} is a discriminative model with a bi-encoder architecture trained on 400 million image-text pairs.
It has one text and one image encoder and is trained to maximize the dot-product of correct image-sentence pairs, and at the same time to minimize the dot-product of random pairs.
We use the pre-trained CLIP model \texttt{RN50x16} for image retrieval, which is the best released CLIP model at the time.\footnote{\url{https://github.com/openai/CLIP/blob/main/clip/clip.py}}

We encode each image $v_k$ in the validation and test sets ($k=20,000$) using CLIP image encoder, and denote each image $v_k$'s 768-dimensional vector representation generated with the image encoder by $\bm{v}_k$.
We similarly encode all \textit{training} English glosses $g_j$ for all nodes in \VisualSem{} using CLIP text encoder, which similarly computes a 768-dimensional vector representation $\bm{g}_j$ for each English gloss.

\begin{table}[t!]
    \centering
    \begin{tabular}{lcccc}
        \toprule
        & {\bf Rank} & \multicolumn{3}{c}{\bf Hits@k}\\
        & {\bf mean (std) $\downarrow$} & \bf 1 $\uparrow$ & \bf 3 $\uparrow$ & \bf 10 $\uparrow$ \\
        \midrule
        $k$-NN & $4,117$ ($16,705$) & 10.0 & 16.5 & 25.6 \\
        \bottomrule
    \end{tabular}
    \caption{Image retrieval results on test images. We report Hits@$k$, which is the percentage of the time the correct node is retrieved in the top-$k$ results (higher is better) and the mean rank of the correct node (lower is better).}
    \label{tab:visualsem_image_retrieval_results}
\end{table}

\paragraph{Evaluation and Discussion}
We use the encoding of each of the $20$k images in the validation and test sets, and frame node retrieval using images as queries as an image-to-gloss ranking problem.
We rank \textit{training} English glosses in \VisualSem{} according to their cosine similarity with the input image, and use the node associated to the gloss with the highest cosine similarity as the retrieved node.
When there are more than one English gloss associated to a node, we use the best ranking across all glosses as the ranking of the retrieved node.
We report results in Table \ref{tab:visualsem_image_retrieval_results}, and note that the quality of the image retrieval module is worse compared to the sentence retrieval.
One of our plans for future work is to investigate how to improve \VisualSem{} image retrieval module.

\section{Related work}
\label{sec:related}

Knowledge bases (KBs) and KGs have a long and rich history, and many publicly available KBs exist and have been built for different purposes.\footnote{We do not differentiate between knowledge bases and knowledge graphs for the purposes of this work and use both terms interchangeably.}
A seminal example is  Cyc \citep{Lenat-etal-1986-CYC}, an early effort towards building a general-purpose knowledge base to store common sense facts and rules about how the world works.
More recent examples of knowledge bases built with different purposes include WordNet \citep{miller1995wordnet,Bond:Paik:2012},
DBPedia \citep{Auer-etal-2007-DBPedia}, Wikidata \citep{Wikidata-2014},
Freebase \citep{Freebase-2008},
YAGO \citep{Rebele2016YAGO,YAGO-2020},
ConceptNet \citep{speer-havasi-2012-representing},
ATOMIC \citep{Sap-etal-2019-ATOMIC},
among many others (see \citealp{kb-survey-2017,ji2020survey} for a detailed list of knowledge bases and algorithms).
Our main goal is to design a KG to support research in vision \& language, which none of the abovementioned KBs are designed to do.

\paragraph{Multi-modal knowledge bases.} 
Two recently proposed multimodal knowledge bases are WN9-IMG \citep{Ruobing-etal-2017-ikrl} and FB15-IMG \citep{moussellySergieh2018multimodal}:
the former consists of a subset of entities and relations from WordNet and was built using the WN18 dataset proposed in \citet{Bordes-etal-2014-wn18}, and additionally includes 10 images illustrating each entity;
the latter is based on the FB15 dataset introduced by \citet{Bordes-etal-2013-freebase}, which in turn consists of examples extracted from Freebase, and also includes 10 images per entity.
Although these KBs include images, they are still restricted to a single data source and constrained in terms of the domains they encompass.

An exception is BabelNet~\citep{navigli2012babelnet}, which is a very large KG that combines varied sources---such as Freebase, WordNet and Wikipedia in several languages, among many others---and includes over $54$ million images mined mostly from Wikipedia and ImageNet.
BabelNet can be seen as a high coverage KB, and in its version v4.0 has more than $15.7$ million concepts in 284 languages.
At times, images linked in BabelNet may have poor quality \citep{calabrese-etal-2020-fatality}, e.g., blurred photos, images loosely related to the concept they illustrate, uninformative images such as icons, flags, or rendered graphs (see Figure \ref{fig:visualsem_example_nodes} for examples).
We build an \textit{image filtering} pipeline that filters out noisy images linked in BabelNet, and we source the remaining high-quality images in \VisualSem{}.

All the aforementioned KBs have in common the fact they are mostly \textit{``entity-centric''}: nodes denote \textit{concepts} and are associated with multimodal information. This is in contrast to ``vision-centric'' KBs such as Visual Genome \citep{krishnavisualgenome-2017} and Visual KB \citep{zhu2015building}, where each instance in the dataset is an image with a \textit{scene graph} representation to support visual query tasks. 
Besides visual queries, multimodal KBs have been designed for conversational reasoning \citep{moon2019opendialkg} and node classification \citep{bloem2021kgbench}.
\VisualSem{} is an entity-centric KG designed to support tasks at the intersection of vision and language.

\section{Conclusions and Future work}
\label{sec:conclusions}

We present the \VisualSem{} knowledge graph with $\sim90$k nodes, over $1.3$M glosses in up to 14 diverse languages, and around $938$k curated and cleaned images illustrating nodes' concepts.
\VisualSem{} bridges a gap in the available resources to train grounded models of language, and is designed to be useful in vision and language research.
We release neural entity retrieval models that accept text and image inputs.
Sentences in any of the $\sim50$ languages supported by multilingual Sentence BERT
can be used to retrieve entities from the KG, and we evaluate retrieval using all 14 languages in \VisualSem{} in Table \ref{tab:sentence_retrieval_results}.
Images can also be used to retrieve nodes using the state-of-the-art CLIP model, and visual entity retrieval is evaluated in Table \ref{tab:visualsem_image_retrieval_results}.
This allows researchers to easily integrate \VisualSem{} into their (neural) model pipelines, and we encourage its use not just in tasks that involve vision and language, but across all sorts of language understanding and/or generation tasks,
in \textit{grounding} and/or in \textit{data augmentation} settings.

\paragraph{Future Work} We will use \VisualSem{} sentence/image retrieval mechanisms and gloss and image features for data augmentation in NLP tasks, e.g. word sense disambiguation and named entity recognition, and on vision and language tasks, e.g. image captioning and visual question answering.
The reason for these tasks is that they all could intuitively benefit from the added knowledge, be it visual, textual, or multi-modal.
We will also investigate how to improve the quality of the image and sentence retrieval modules, both central to the KG impact on current neural-based models.
Finally, we plan to improve and grow the KG, which is under active development. We will particularly gauge whether there is interest from the research community in increasing its coverage to include other parts of, or the entirety of, Wikipedia, for example.

\section*{Acknowledgements}
IC has received funding from the European Union’s Horizon 2020 research and innovation programme under the Marie Sk\l{}odowska-Curie grant agreement No 838188. KC is partly supported by Samsung Advanced Institute of Technology (Next Generation Deep Learning: from pattern recognition to AI) and Samsung Research (Improving Deep Learning using Latent Structure). KC also thanks Naver, eBay, NVIDIA, and NSF Award 1922658 for support. CV's work on this project at New York University was financially supported by Eric and Wendy Schmidt (made by recommendation of the Schmidt Futures program) and Samsung Research (under the project \textit{Improving Deep Learning using Latent Structure}) and benefitted from in-kind support by the NYU High-Performance Computing Center. This material is based upon work supported by the National Science Foundation under Grant No. 1922658. Any opinions, findings, and conclusions or recommendations expressed in this material are those of the author(s) and do not necessarily reflect the views of the National Science Foundation.

\bibliography{anthology,refs}

\begin{thebibliography}{38}
\expandafter\ifx\csname natexlab\endcsname\relax\def\natexlab#1{#1}\fi

\bibitem[{Alberts and Calixto(2020)}]{ImagiFilter2020}
Houda Alberts and Iacer Calixto. 2020.
\newblock \href {http://arxiv.org/} {Imagifilter: A resource to enable the
  semi-automatic mining of images at scale.}
\newblock \emph{arXiv preprint}.

\bibitem[{Auer et~al.(2007)Auer, Bizer, Kobilarov, Lehmann, Cyganiak, and
  Ives}]{Auer-etal-2007-DBPedia}
S{\"o}ren Auer, Christian Bizer, Georgi Kobilarov, Jens Lehmann, Richard
  Cyganiak, and Zachary Ives. 2007.
\newblock Dbpedia: A nucleus for a web of open data.
\newblock In \emph{The Semantic Web}, pages 722--735, Berlin, Heidelberg.
  Springer Berlin Heidelberg.

\bibitem[{Blei et~al.(2003)Blei, Ng, and Jordan}]{Blei-etal-2003-LDA}
David~M. Blei, Andrew~Y. Ng, and Michael~I. Jordan. 2003.
\newblock Latent dirichlet allocation.
\newblock \emph{J. Mach. Learn. Res.}, 3(null):993–1022.

\bibitem[{Bloem et~al.(2021)Bloem, Wilcke, van Berkel, and
  de~Boer}]{bloem2021kgbench}
Peter Bloem, Xander Wilcke, Lucas van Berkel, and Victor de~Boer. 2021.
\newblock \href {https://openreview.net/forum?id=yeK_9wxRDbA} {kgbench: A
  collection of knowledge graph datasets for evaluating relational and
  multimodal machine learning}.
\newblock In \emph{Eighteenth Extended Semantic Web Conference - Resources
  Track}.

\bibitem[{Bojanowski et~al.(2016)Bojanowski, Grave, Joulin, and
  Mikolov}]{bojanowski2016enriching}
Piotr Bojanowski, Edouard Grave, Armand Joulin, and Tomas Mikolov. 2016.
\newblock Enriching word vectors with subword information.
\newblock \emph{arXiv preprint arXiv:1607.04606}.

\bibitem[{Bollacker et~al.(2008)Bollacker, Evans, Paritosh, Sturge, and
  Taylor}]{Freebase-2008}
Kurt Bollacker, Colin Evans, Praveen Paritosh, Tim Sturge, and Jamie Taylor.
  2008.
\newblock \href {https://doi.org/10.1145/1376616.1376746} {Freebase: A
  collaboratively created graph database for structuring human knowledge}.
\newblock In \emph{Proceedings of the 2008 ACM SIGMOD International Conference
  on Management of Data}, SIGMOD ’08, page 1247–1250, New York, NY, USA.
  Association for Computing Machinery.

\bibitem[{Bond and Paik(2012)}]{Bond:Paik:2012}
Francis Bond and Kyonghee Paik. 2012.
\newblock A survey of wordnets and their licenses.
\newblock In \emph{Proceedings of the 6th Global WordNet Conference (GWC
  2012)}, Matsue.
\newblock 64--71.

\bibitem[{Bordes et~al.(2014)Bordes, Glorot, Weston, and
  Bengio}]{Bordes-etal-2014-wn18}
Antoine Bordes, Xavier Glorot, Jason Weston, and Yoshua Bengio. 2014.
\newblock \href {https://doi.org/10.1007/s10994-013-5363-6} {A semantic
  matching energy function for learning with multi-relational data}.
\newblock \emph{Mach. Learn.}, 94(2):233–259.

\bibitem[{Bordes et~al.(2013)Bordes, Usunier, Garcia-Dur\'{a}n, Weston, and
  Yakhnenko}]{Bordes-etal-2013-freebase}
Antoine Bordes, Nicolas Usunier, Alberto Garcia-Dur\'{a}n, Jason Weston, and
  Oksana Yakhnenko. 2013.
\newblock Translating embeddings for modeling multi-relational data.
\newblock In \emph{Proceedings of the 26th International Conference on Neural
  Information Processing Systems - Volume 2}, NIPS’13, page 2787–2795, Red
  Hook, NY, USA. Curran Associates Inc.

\bibitem[{Calabrese et~al.(2020)Calabrese, Bevilacqua, and
  Navigli}]{calabrese-etal-2020-fatality}
Agostina Calabrese, Michele Bevilacqua, and Roberto Navigli. 2020.
\newblock \href {https://doi.org/10.18653/v1/2020.acl-main.425} {Fatality
  killed the cat or: {B}abel{P}ic, a multimodal dataset for non-concrete
  concepts}.
\newblock In \emph{Proceedings of the 58th Annual Meeting of the Association
  for Computational Linguistics}, pages 4680--4686, Online. Association for
  Computational Linguistics.

\bibitem[{Colla et~al.(2018)Colla, Mensa, Radicioni, and Lieto}]{Collaetal2018}
Davide Colla, Enrico Mensa, Daniele~P. Radicioni, and Antonio Lieto. 2018.
\newblock Tell me why: Computational explanation of conceptual similarity
  judgments.
\newblock In \emph{Information Processing and Management of Uncertainty in
  Knowledge-Based Systems. Theory and Foundations}, pages 74--85, Cham.
  Springer International Publishing.

\bibitem[{{Cui} et~al.(2018){Cui}, {Liu}, and {Zhu}}]{Cuietal2018}
P.~{Cui}, S.~{Liu}, and W.~{Zhu}. 2018.
\newblock General knowledge embedded image representation learning.
\newblock \emph{IEEE Transactions on Multimedia}, 20(1):198--207.

\bibitem[{{Deng} et~al.(2009){Deng}, {Dong}, {Socher}, {Li}, {Kai Li}, and {Li
  Fei-Fei}}]{Deng-etal-2009-ImageNet}
J.~{Deng}, W.~{Dong}, R.~{Socher}, L.~{Li}, {Kai Li}, and {Li Fei-Fei}. 2009.
\newblock Imagenet: A large-scale hierarchical image database.
\newblock In \emph{2009 IEEE Conference on Computer Vision and Pattern
  Recognition}, pages 248--255.

\bibitem[{Dieng et~al.(2020)Dieng, Ruiz, and Blei}]{dieng-etal-2020-topic}
Adji~B. Dieng, Francisco J.~R. Ruiz, and David~M. Blei. 2020.
\newblock \href {https://doi.org/10.1162/tacl_a_00325} {Topic modeling in
  embedding spaces}.
\newblock \emph{Transactions of the Association for Computational Linguistics},
  8:439--453.

\bibitem[{Eastlake and Jones(2001)}]{eastlake-jones:2001sha1}
D.~Eastlake and P.~Jones. 2001.
\newblock Rfc3174: Us secure hash algorithm 1 (sha1).

\bibitem[{Guu et~al.(2020)Guu, Lee, Tung, Pasupat, and
  Chang}]{pmlr-v119-guu20a}
Kelvin Guu, Kenton Lee, Zora Tung, Panupong Pasupat, and Mingwei Chang. 2020.
\newblock \href {http://proceedings.mlr.press/v119/guu20a.html} {Realm:
  Retrieval-augmented language model pre-training}.
\newblock In \emph{Proceedings of the 37th International Conference on Machine
  Learning}, volume 119 of \emph{Proceedings of Machine Learning Research},
  pages 3929--3938. PMLR.

\bibitem[{Ji et~al.(2020)Ji, Pan, Cambria, Marttinen, and Yu}]{ji2020survey}
Shaoxiong Ji, Shirui Pan, Erik Cambria, Pekka Marttinen, and Philip~S Yu. 2020.
\newblock A survey on knowledge graphs: Representation, acquisition and
  applications.
\newblock \emph{arXiv preprint arXiv:2002.00388}.

\bibitem[{Krishna et~al.(2017)Krishna, Zhu, Groth, Johnson, Hata, Kravitz,
  Chen, Kalantidis, Li, Shamma, and et~al.}]{krishnavisualgenome-2017}
Ranjay Krishna, Yuke Zhu, Oliver Groth, Justin Johnson, Kenji Hata, Joshua
  Kravitz, Stephanie Chen, Yannis Kalantidis, Li-Jia Li, David~A. Shamma, and
  et~al. 2017.
\newblock \href {https://doi.org/10.1007/s11263-016-0981-7} {Visual genome:
  Connecting language and vision using crowdsourced dense image annotations}.
\newblock \emph{International Journal of Computer Vision}, 123(1):32–73.

\bibitem[{Lenat et~al.(1986)Lenat, Prakash, and Shepherd}]{Lenat-etal-1986-CYC}
Doug Lenat, Mayank Prakash, and Mary Shepherd. 1986.
\newblock Cyc: Using common sense knowledge to overcome brittleness and
  knowledge acquistion bottlenecks.
\newblock \emph{AI Mag.}, 6(4):65–85.

\bibitem[{Lewis et~al.(2020{\natexlab{a}})Lewis, Ghazvininejad, Ghosh,
  Aghajanyan, Wang, and Zettlemoyer}]{NEURIPS2020-MARGE}
Mike Lewis, Marjan Ghazvininejad, Gargi Ghosh, Armen Aghajanyan, Sida Wang, and
  Luke Zettlemoyer. 2020{\natexlab{a}}.
\newblock \href
  {https://proceedings.neurips.cc/paper/2020/file/d6f1dd034aabde7657e6680444ceff62-Paper.pdf}
  {Pre-training via paraphrasing}.
\newblock In \emph{Advances in Neural Information Processing Systems},
  volume~33, pages 18470--18481. Curran Associates, Inc.

\bibitem[{Lewis et~al.(2020{\natexlab{b}})Lewis, Perez, Piktus, Petroni,
  Karpukhin, Goyal, K\"{u}ttler, Lewis, Yih, Rockt\"{a}schel, Riedel, and
  Kiela}]{NEURIPS2020-RAG}
Patrick Lewis, Ethan Perez, Aleksandra Piktus, Fabio Petroni, Vladimir
  Karpukhin, Naman Goyal, Heinrich K\"{u}ttler, Mike Lewis, Wen-tau Yih, Tim
  Rockt\"{a}schel, Sebastian Riedel, and Douwe Kiela. 2020{\natexlab{b}}.
\newblock \href
  {https://proceedings.neurips.cc/paper/2020/file/6b493230205f780e1bc26945df7481e5-Paper.pdf}
  {Retrieval-augmented generation for knowledge-intensive nlp tasks}.
\newblock In \emph{Advances in Neural Information Processing Systems},
  volume~33, pages 9459--9474. Curran Associates, Inc.

\bibitem[{van~der Maaten and Hinton(2008)}]{vanDerMaaten2008}
Laurens van~der Maaten and Geoffrey Hinton. 2008.
\newblock \href {http://www.jmlr.org/papers/v9/vandermaaten08a.html}
  {Visualizing data using {t-SNE}}.
\newblock \emph{Journal of Machine Learning Research}, 9:2579--2605.

\bibitem[{Miller(1995)}]{miller1995wordnet}
George~A Miller. 1995.
\newblock Wordnet: a lexical database for english.
\newblock \emph{Communications of the ACM}, 38(11):39--41.

\bibitem[{Moon et~al.(2019)Moon, Shah, Kumar, and Subba}]{moon2019opendialkg}
Seungwhan Moon, Pararth Shah, Anuj Kumar, and Rajen Subba. 2019.
\newblock Opendialkg: Explainable conversational reasoning with attention-based
  walks over knowledge graphs.
\newblock In \emph{Proceedings of the 57th Annual Meeting of the Association
  for Computational Linguistics}, pages 845--854.

\bibitem[{Mousselly~Sergieh et~al.(2018)Mousselly~Sergieh, Botschen, Gurevych,
  and Roth}]{moussellySergieh2018multimodal}
Hatem Mousselly~Sergieh, Teresa Botschen, Iryna Gurevych, and Stefan Roth.
  2018.
\newblock {A Multimodal Translation-Based Approach for Knowledge Graph
  Representation Learning}.
\newblock In \emph{Proceedings of the 7th Joint Conference on Lexical and
  Computational Semantics (*SEM 2018)}, page to appear. Association for
  Computational Linguistics.

\bibitem[{Navigli and Ponzetto(2012)}]{navigli2012babelnet}
Roberto Navigli and Simone~Paolo Ponzetto. 2012.
\newblock Babelnet: The automatic construction, evaluation and application of a
  wide-coverage multilingual semantic network.
\newblock \emph{Artificial Intelligence}, 193:217--250.

\bibitem[{Pellissier~Tanon et~al.(2020)Pellissier~Tanon, Weikum, and
  Suchanek}]{YAGO-2020}
Thomas Pellissier~Tanon, Gerhard Weikum, and Fabian Suchanek. 2020.
\newblock Yago 4: A reason-able knowledge base.
\newblock In \emph{The Semantic Web}, pages 583--596, Cham. Springer
  International Publishing.

\bibitem[{Radford et~al.(2021)Radford, Kim, Hallacy, Ramesh, Goh, Agarwal,
  Sastry, Askell, Mishkin, Clark et~al.}]{radford2021learning}
Alec Radford, Jong~Wook Kim, Chris Hallacy, Aditya Ramesh, Gabriel Goh,
  Sandhini Agarwal, Girish Sastry, Amanda Askell, Pamela Mishkin, Jack Clark,
  et~al. 2021.
\newblock Learning transferable visual models from natural language
  supervision.
\newblock \emph{arXiv preprint arXiv:2103.00020}.

\bibitem[{Rebele et~al.(2016)Rebele, Suchanek, Hoffart, Biega, Kuzey, and
  Weikum}]{Rebele2016YAGO}
Thomas Rebele, Fabian~M. Suchanek, Johannes Hoffart, Joanna~Asia Biega, Erdal
  Kuzey, and Gerhard Weikum. 2016.
\newblock Yago: A multilingual knowledge base from wikipedia, wordnet, and
  geonames.
\newblock In \emph{International Semantic Web Conference}.

\bibitem[{Reimers and Gurevych(2019)}]{reimers-2019-sentence-bert}
Nils Reimers and Iryna Gurevych. 2019.
\newblock \href {http://arxiv.org/abs/1908.10084} {Sentence-bert: Sentence
  embeddings using siamese bert-networks}.
\newblock In \emph{Proceedings of the 2019 Conference on Empirical Methods in
  Natural Language Processing}. Association for Computational Linguistics.

\bibitem[{Reimers and Gurevych(2020)}]{reimers-2020-multilingual-sentence-bert}
Nils Reimers and Iryna Gurevych. 2020.
\newblock \href {http://arxiv.org/abs/2004.09813} {Making monolingual sentence
  embeddings multilingual using knowledge distillation}.
\newblock \emph{arXiv preprint arXiv:2004.09813}.

\bibitem[{Russakovsky et~al.(2015)Russakovsky, Deng, Su, Krause, Satheesh, Ma,
  Huang, Karpathy, Khosla, Bernstein, Berg, and Fei-Fei}]{ILSVRC2015}
Olga Russakovsky, Jia Deng, Hao Su, Jonathan Krause, Sanjeev Satheesh, Sean Ma,
  Zhiheng Huang, Andrej Karpathy, Aditya Khosla, Michael Bernstein,
  Alexander~C. Berg, and Li~Fei-Fei. 2015.
\newblock \href {https://doi.org/10.1007/s11263-015-0816-y} {Imagenet large
  scale visual recognition challenge}.
\newblock \emph{Int. J. Comput. Vision}, 115(3):211–252.

\bibitem[{Sap et~al.(2019)Sap, Bras, Allaway, Bhagavatula, Lourie, Rashkin,
  Roof, Smith, and Choi}]{Sap-etal-2019-ATOMIC}
Maarten Sap, Ronan~Le Bras, Emily Allaway, Chandra Bhagavatula, Nicholas
  Lourie, Hannah Rashkin, Brendan Roof, Noah~A. Smith, and Yejin Choi. 2019.
\newblock \href {https://doi.org/10.1609/aaai.v33i01.33013027} {{ATOMIC:} an
  atlas of machine commonsense for if-then reasoning}.
\newblock In \emph{The Thirty-Third {AAAI} Conference on Artificial
  Intelligence, {AAAI} 2019}, pages 3027--3035. {AAAI} Press.

\bibitem[{Speer and Havasi(2012)}]{speer-havasi-2012-representing}
Robyn Speer and Catherine Havasi. 2012.
\newblock \href
  {http://www.lrec-conf.org/proceedings/lrec2012/pdf/1072_Paper.pdf}
  {Representing general relational knowledge in {C}oncept{N}et 5}.
\newblock In \emph{Proceedings of the Eighth International Conference on
  Language Resources and Evaluation ({LREC}-2012)}, pages 3679--3686, Istanbul,
  Turkey. European Languages Resources Association (ELRA).

\bibitem[{Vrande\v{c}i\'{c} and Kr\"{o}tzsch(2014)}]{Wikidata-2014}
Denny Vrande\v{c}i\'{c} and Markus Kr\"{o}tzsch. 2014.
\newblock \href {https://doi.org/10.1145/2629489} {Wikidata: A free
  collaborative knowledgebase}.
\newblock \emph{Commun. ACM}, 57(10):78–85.

\bibitem[{{Wang} et~al.(2017){Wang}, {Mao}, {Wang}, and {Guo}}]{kb-survey-2017}
Q.~{Wang}, Z.~{Mao}, B.~{Wang}, and L.~{Guo}. 2017.
\newblock Knowledge graph embedding: A survey of approaches and applications.
\newblock \emph{IEEE Transactions on Knowledge and Data Engineering},
  29(12):2724--2743.

\bibitem[{Xie et~al.(2017)Xie, Liu, Luan, and Sun}]{Ruobing-etal-2017-ikrl}
Ruobing Xie, Zhiyuan Liu, Huanbo Luan, and Maosong Sun. 2017.
\newblock Image-embodied knowledge representation learning.
\newblock In \emph{Proceedings of the 26th International Joint Conference on
  Artificial Intelligence}, IJCAI’17, page 3140–3146. AAAI Press.

\bibitem[{Zhu et~al.(2015)Zhu, Zhang, R{\'e}, and Fei-Fei}]{zhu2015building}
Yuke Zhu, Ce~Zhang, Christopher R{\'e}, and Li~Fei-Fei. 2015.
\newblock Building a large-scale multimodal knowledge base system for answering
  visual queries.
\newblock \emph{arXiv preprint arXiv:1507.05670}.

\end{thebibliography}
\bibliographystyle{acl_natbib}

\clearpage
\appendix
\section{Examples}\label{sec:appendix_examples}
\begin{figure*}[ht!]
    \centering
    \includegraphics[width=.65\textwidth]{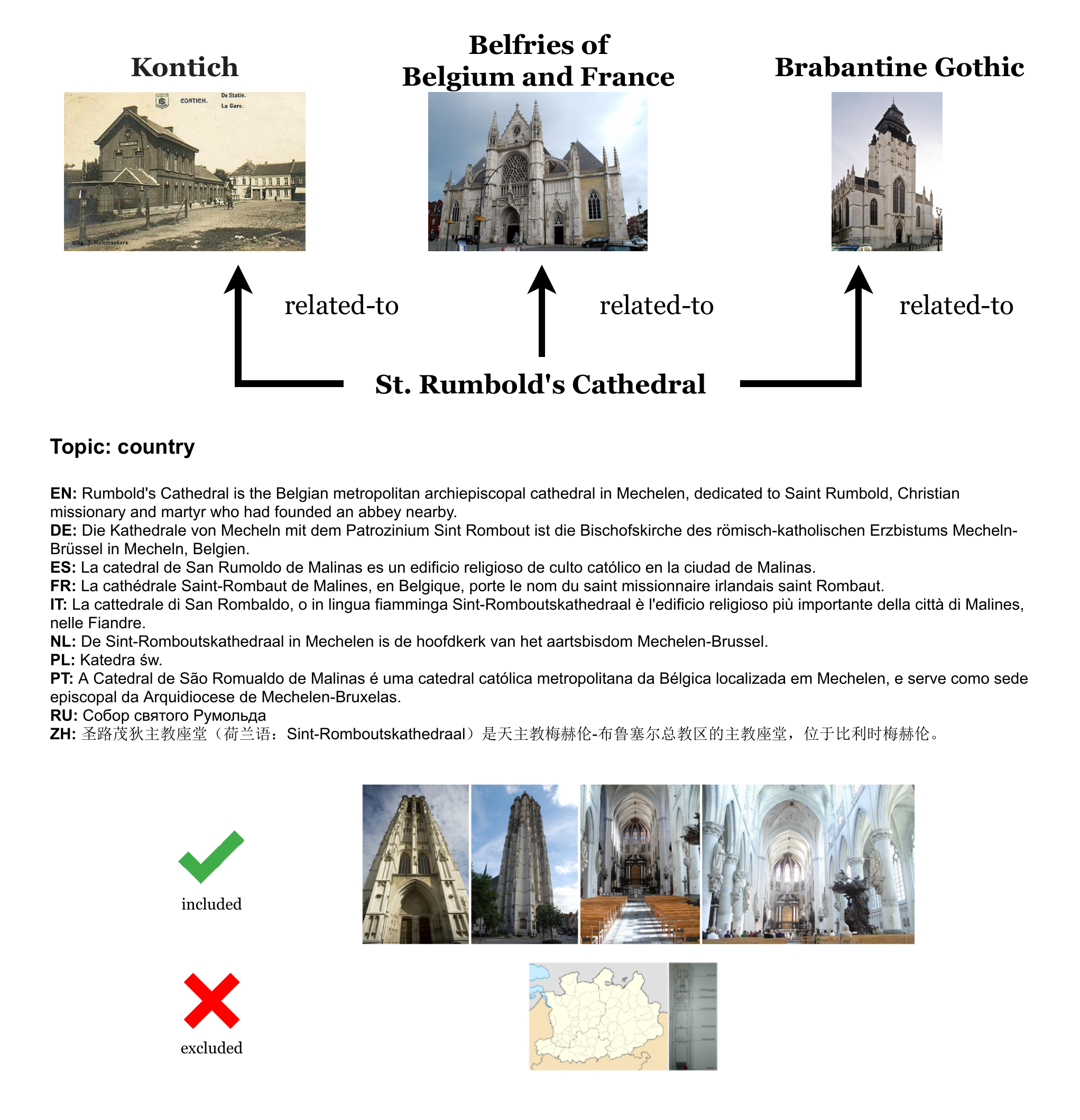}
    \caption{\textit{St. Rumbold's Cathedral}. The node is connected to four other nodes (3 shown) with relation types \texttt{related-to} and \texttt{has-a} (not shown), and has a total of 23 images kept (4 shown).}
    \label{fig:visualsem_example_st-rumbolds-cathedral}
\end{figure*}

\begin{figure*}[ht!]
    \centering
    \includegraphics[width=.65\textwidth]{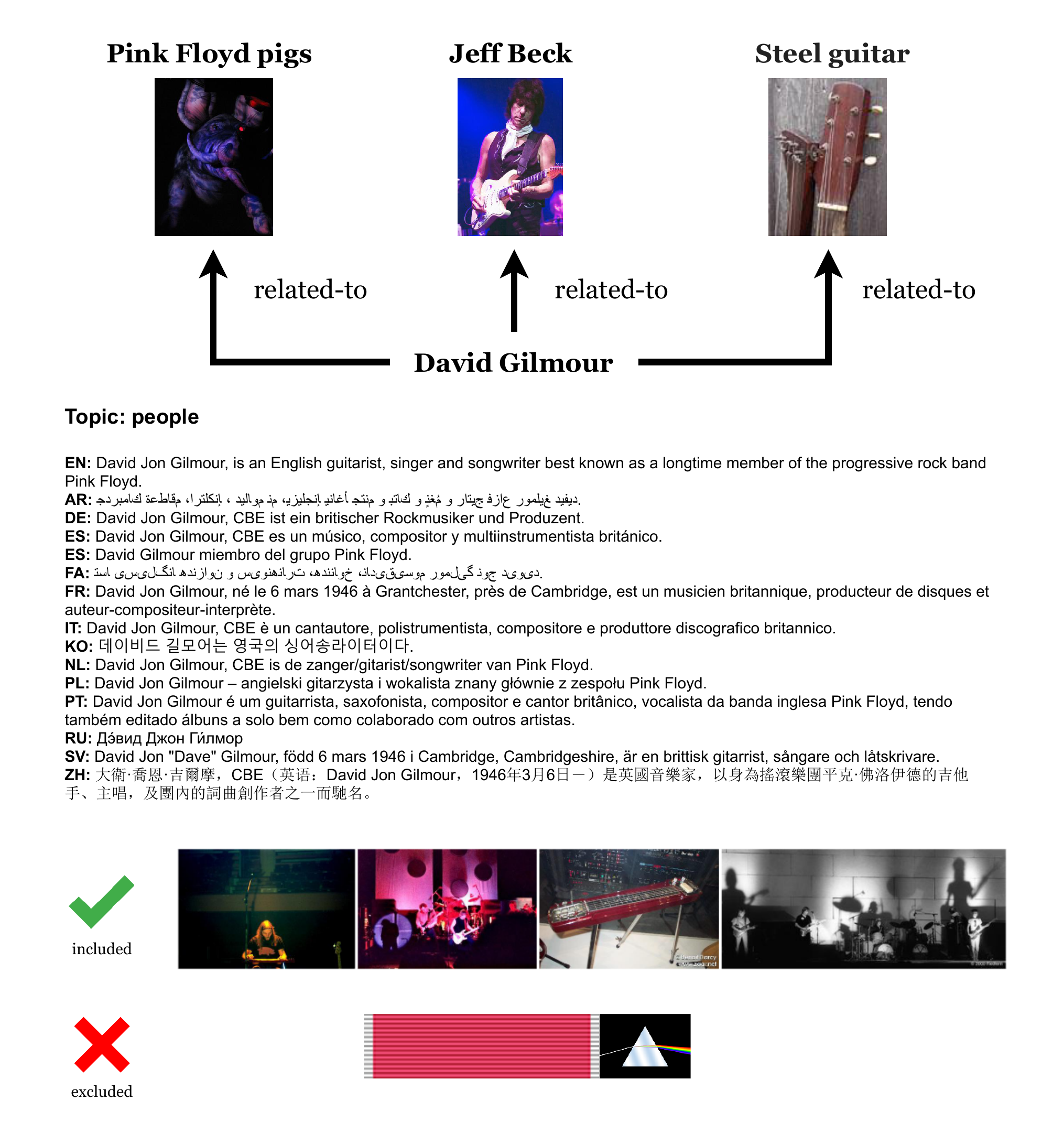}
    \caption{\textit{David Gilmour}. The node is connected to 10 other nodes (3 shown) with 2 relation types \texttt{related-to} and \texttt{has-a} (not shown), and has a total of 10 images kept (4 shown).}
    \label{fig:visualsem_example_david-gilmour}
\end{figure*}

\begin{figure*}[ht!]
    \centering
    \includegraphics[width=.6\textwidth]{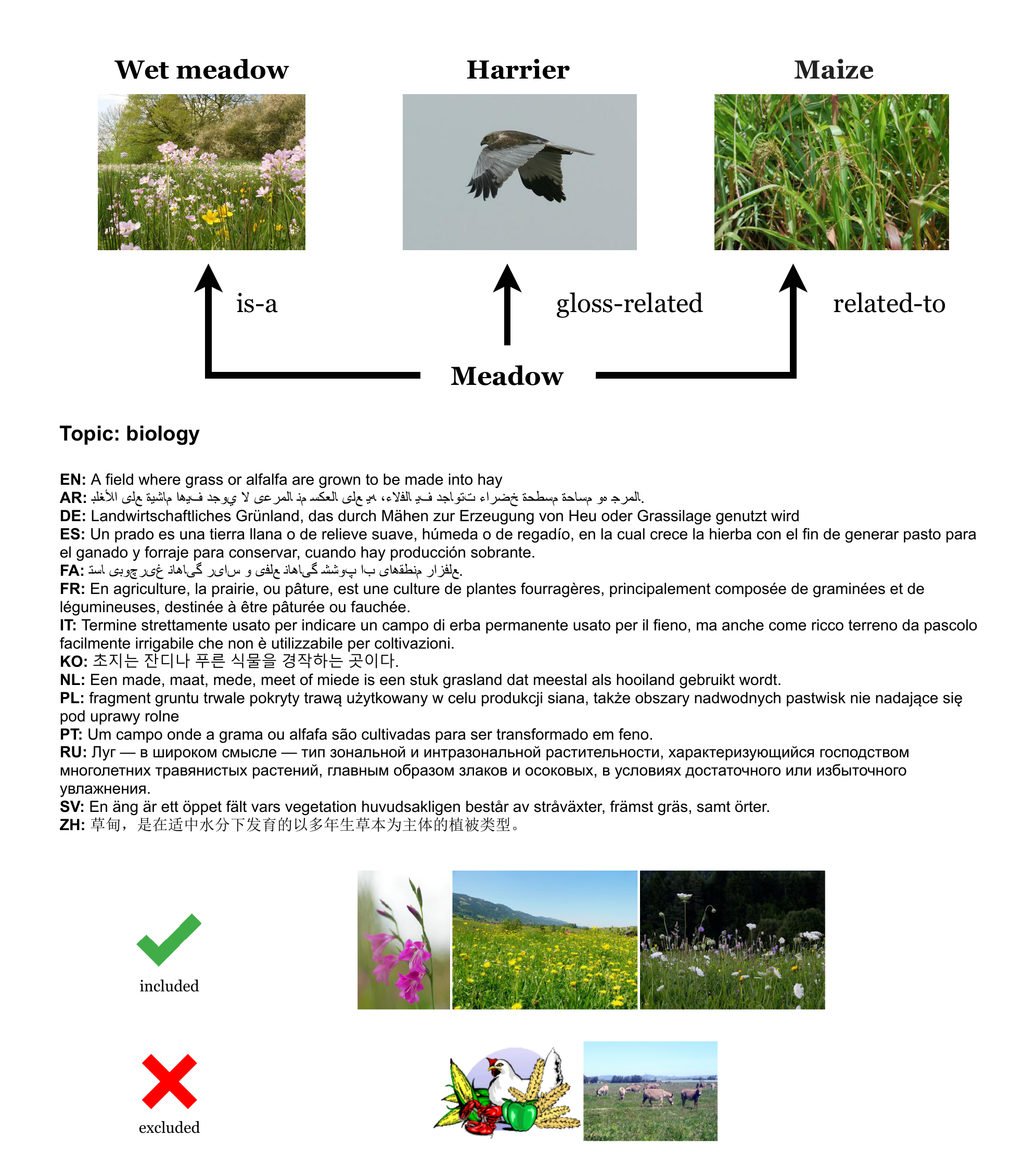}
    \caption{\textit{Meadow}. The node is connected to 127 other nodes (3 shown) with relation types \texttt{related-to}, \texttt{gloss-related}, \texttt{is-a}, \texttt{has-a} (not shown), and has a total of 26 images kept (3 shown).}
    \label{fig:visualsem_example_meadow}
\end{figure*}

\begin{figure*}[ht!]
    \centering
    \includegraphics[width=.6\textwidth]{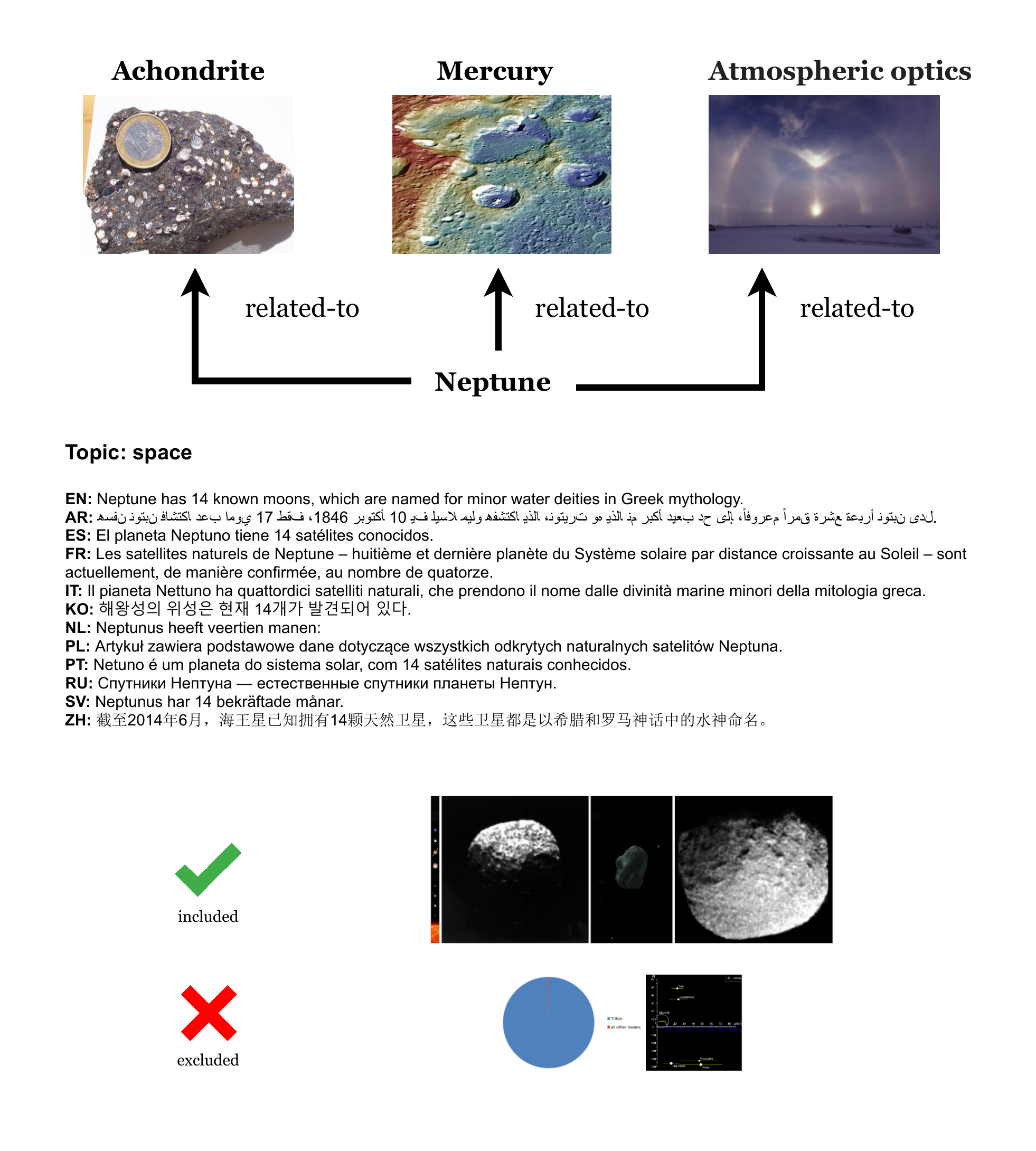}
    \caption{\textit{Neptune}. The node is connected to 10 other nodes (3 shown) with relation types \texttt{related-to}, \texttt{is-a} (not shown), \texttt{has-a} (not shown), and has a total of 26 images kept (4 shown).}
    \label{fig:visualsem_example_neptune}
\end{figure*}

\begin{figure*}[ht!]
    \centering
    \includegraphics[width=.85\textwidth]{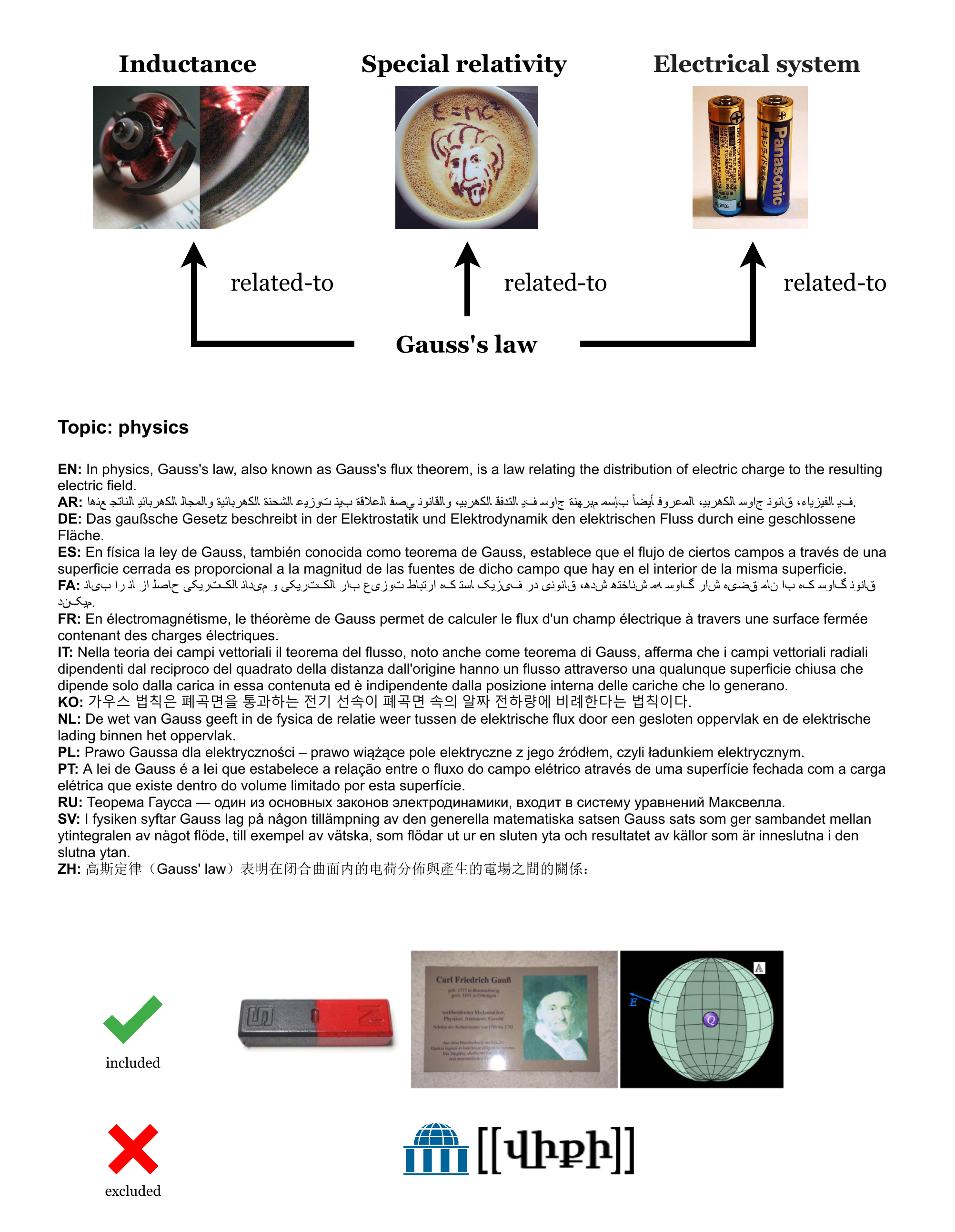}
    \caption{\textit{Gauss' Law}. The node is connected to 6 other nodes (3 shown) with relation type \texttt{related-to}, and has a total of 5 images kept (3 shown).}
    \label{fig:visualsem_example_gausslaw}
\end{figure*}

We now show some example data in \VisualSem{} to illustrate what kind of information is available in a node, including metadata, as well as what images are kept and what are filtered out.
We select five topics (i.e., country, people, physics, biology, space) in the topic model discussed in Section~\ref{sec:topic_models} and show one example node in each, including:
{\bf multilingual glosses} in up to 14 languages (we show a maximum of one gloss per language per node to avoid clutter);
{\bf images} linked to the node, including the ones filtered out by our procedure illustrated in Section~\ref{sec:approach};
{\bf relations} to other nodes in the KG (we show a maximum of three connected nodes to avoid clutter).
We illustrate the five node examples in Figures \ref{fig:visualsem_example_st-rumbolds-cathedral}--\ref{fig:visualsem_example_gausslaw}, and we use these examples to support a discussion on the quality of the data.

First, we note that information described in a node's glosses can be repetitive across languages (e.g., in Figure \ref{fig:visualsem_example_meadow} the glosses in English and Portuguese are translations of one another), but often can also be complementary (e.g., again in Figure \ref{fig:visualsem_example_meadow} the gloss in French includes novel information such as the mention to leguminous crops, which does not appear neither in English nor Portuguese).

We also note that the removed images are most of the time non-photographic and/or noisy, in accord with our motivation. Images such as the map or the sketch in Figure \ref{fig:visualsem_example_st-rumbolds-cathedral}, or the excluded images in Figures \ref{fig:visualsem_example_neptune} or \ref{fig:visualsem_example_gausslaw}, are not \textit{descriptive} of the node.
However, a limited number of relevant images are sometimes excluded by our procedure (e.g., the field with animals in Figure \ref{fig:visualsem_example_meadow}), and conversely sometimes images that are tangentially related to a node are kept (e.g., the magnet in Figure \ref{fig:visualsem_example_gausslaw}).

\end{document}